\documentclass{article} 
\usepackage{iclr2021_conference,times}

\usepackage{amsmath,amsfonts,bm}

\def\eqref#1{equation~\ref{#1}}

\def\1{\bm{1}}

\DeclareMathAlphabet{\mathsfit}{\encodingdefault}{\sfdefault}{m}{sl}
\SetMathAlphabet{\mathsfit}{bold}{\encodingdefault}{\sfdefault}{bx}{n}

\usepackage{hyperref}
\usepackage{url}
\usepackage{graphicx}
\usepackage{wrapfig}
\usepackage{multirow}
\usepackage{booktabs}
\usepackage{lipsum}

\title{Causal-TGAN: Generating Tabular Data \\ Using Causal Generative Adversarial Networks}

\author{Bingyang Wen, Luis Oliveros Colon, K.P. Subbalakshmi \& R. Chandramouli \\
Department of Electrical and Computer Engineering\\
Stevens Institute of Technology\\
Hoboken, NJ 07030, USA \\
\texttt{bwen4@stevens.edu} \\
}

\iclrfinalcopy 
\begin{document}

\maketitle

\begin{abstract}
Synthetic data generation becomes prevalent as a solution to privacy leakage and data shortage. Generative models are designed to generate a realistic synthetic dataset, which can precisely express the data distribution for the real dataset. The generative adversarial networks (GAN), which gain great success in the computer vision fields, are doubtlessly used for synthetic data generation.  Though there are prior works that have demonstrated great progress, most of them learn the correlations in the data distributions rather than the true processes in which the datasets are naturally generated. Correlation is not reliable for it is a statistical technique that only tells linear dependencies and is easily affected by the dataset's bias. Causality, which encodes all underlying factors of how the real data be naturally generated, is more reliable than correlation. In this work, we propose a causal model named Causal Tabular Generative Neural Network (Causal-TGAN) to generate synthetic tabular data using the tabular data's causal information. Extensive experiments on both simulated datasets and real datasets demonstrate better performance of our method when given the true causal graph and a comparable performance when using the estimated causal graph.

\end{abstract}

\section{Introduction \& Related Work}
\vspace{-0.2cm}


Over the last few years, there has been an increase in the work done around GANs (\cite{Goodfellow14}). This architecture has been used to generate realistic images with great success in the field of computer vision. More recently, these generative models have been used for tabular data generation, demonstrating better performance than traditional generative models based on statistical techniques. GANs offer more flexibility in modeling real data distributions and provide higher quality results. 

Prior to this work, \cite{Camino18} propose to generate tabular data using GANs by introducing a multi-categorical setting. \cite{Park18} present TableGAN, which can synthesize data containing categorical, continuous, and discrete values. The model adopts DCGAN architecture and guides the generator to generate synthetic samples with a reasonable label by adding an auxiliary classifier. Motivated by the urge to keep the data's privacy, \cite{Jordon19} propose PATE-GAN to generate differentially private synthetic data, especially suitable for sensitive data such as Electronic Health Records (EHR). With the same motivations, \cite{Torfi20} present CorGAN, which uses a one-dimension convolutional GAN architecture for both generator and discriminator. Moreover, \cite{Torfi20} employ an autoencoder to transform the continuous outputs into discrete values. CTGAN (\cite{Xu19}) addresses class imbalance by introducing a conditional generator, which provides the model with the capacity to evenly re-sampling all categories from discrete columns during the training process. In addition, CTGAN employs the Gaussian Mixture Models to encode the continuous columns to adapt the multimodal distributions for these columns better. This paper also introduces a new benchmark, looking to establish an accurate comparison between the different models that use GANs for generating synthetic tabular data.

However, all the GANs mentioned above learn the target distribution by capturing the inter-variants correlations. Excessive reliance on the inter-variants correlations may limit the performance of using GANs to generate tabular data. It is a fact that the all the real dataset is created from the underlying inter-variants causal relations rather than the correlations. Using GANs to simulate how the real dataset is created in nature is supposed to be more efficient than simply learning the data distributions.
We hence propose Causal Tabular Generative Adversarial Network (Causal-TGAN) to take advantage of inter-variants causal relations. We use adversarial learning to train a structural causal model (SCM) (\cite{pearl09}). Similar methods have been used for causal discovery (\cite{gouEtal18}) and intervened image attributes generation (\cite{kocEtal17}) in previous works. To our best knowledge, we are the first to use the adversarially trained SCM for synthetic data generation. Moreover, compared with the aforementioned methods, which are designed only to generate continuous variables (\cite{gouEtal18}) or binary variables (\cite{kocEtal17}), our Causal-TGAN can generate more types of variables such as categorical and ordinal.  

We use two kinds of metrics to evaluate Causal-TGAN on both the simulated dataset and real dataset. The experimental results demonstrate the advantage of Causal-TGAN compared with other GANs when given the true causal relations. When the true causal relations are absent, our Causal-TGAN still shows comparative performance using the estimated causal relations.

\vspace{-0.2cm}
\section{Preliminary}
\vspace{-0.3cm}
\label{sec:preliminary}
\begin{wrapfigure}{R}{0.35\textwidth}
\centering
\includegraphics[width=0.35\textwidth]{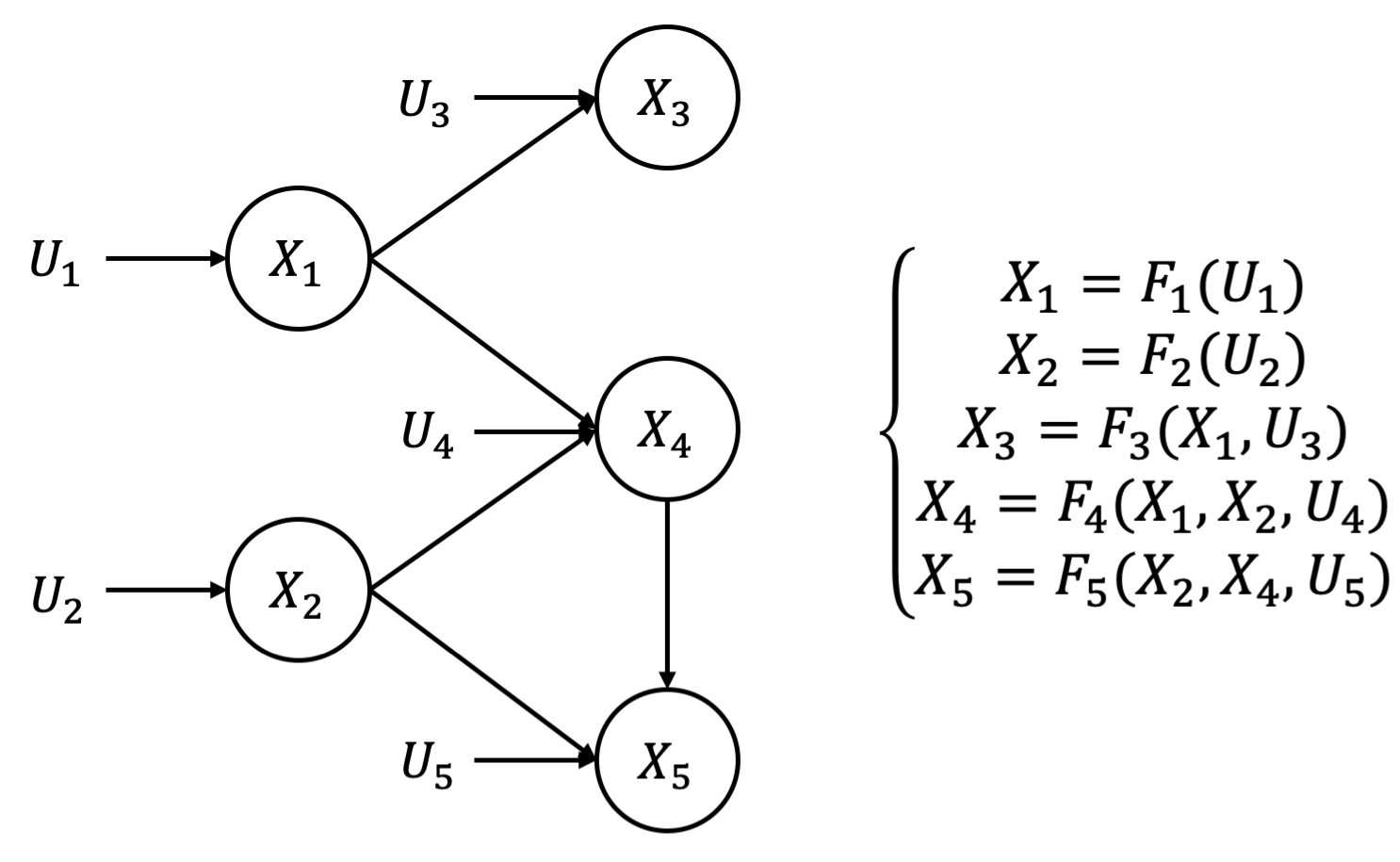}
\caption{\label{fig:scm_example} \textbf{An example of SCM:} The causal graph is shown in the left diagram and all causal mechanisms are shown in the right equations.}
\end{wrapfigure}
In this section, we briefly introduce the concepts of causal models, specifically, the structural causal models (SCM) (\cite{pearl09}). Let $\mathcal A$ be the set of features on a dataset $D$. $D$ can be regarded as a joint distribution over $\mathcal A$'s domain. A causal model on $\mathcal A$ is a way to compactly describe this joint distribution by specifying the causal relations among the certain features in $\mathcal A$. Specifically, an SCM is a directed acyclic graph (DAG) in which each feature of $\mathcal A$ is represented as a node, and the causal relations are represented by the causal mechanisms. 

Formally, a SCM, $\mathcal M$, can be represented by a triplet $\mathcal M = \langle X, \mathcal F, \mathcal U \rangle$ that contains a set of endogenous variables $ X=\{X_1, X_2,..., X_d\}$, a set of causal mechanisms $\mathcal F=\{F_1, F_2,..., F_d\}$, and a set of exogenous variables $U=\{U_1, U_2,..., U_d\}$, where each $U_i$ is independently drawn after distribution $\mathcal U$. In our case, each endogenous variable $X$ represents a feature in $\mathcal A$. Under the causal sufficiency assumption (i.e., no confounder), the causal relation of ``$X_i$ causes $X_j$'' can be represented by: $X_j = F_{j}(X_i, U_{j})$. $X_i$ is called parent node or direct cause of $X_j$ and the exogenous variable $U_{j}$ can be seen as the sum of all unobserved causes of $X_j$. The causal graph $\mathcal G_{\mathcal M}$ is a path diagram that encodes all of these causal relations for $\mathcal M$. Figure \ref{fig:scm_example} illustrate an example of SCM.

\vspace{-0.2cm}
\section{Method}
\vspace{-0.3cm}
Causal-TGAN is an SCM that employs multiple neural networks to fit the causal mechanisms for a causal graph using adversarial learning.
Causal-TGAN runs in two steps: (i) obtain the causal graph that encodes the underlying inter-variant's causal relations of the target dataset. (ii) Using the obtained causal graph to construct Causal-TGAN and then adversarially training the Causal-TGAN to generate realistic samples. We discuss in details the above steps in this section. In Section \ref{sec:causal_graph}, we introduce the method we use for discovering a causal graph from the tabular data when the ground causal graph is absent. In Section \ref{sec:causal_TGAN}, we introduce the construction of Causal-TGAN as an SCM given a causal graph and introduce the training objective. 
\vspace{-0.2cm}
\subsection{Causal Discovery}
\vspace{-0.2cm}
\label{sec:causal_graph}
We use PC algorithm (\cite{pc00}) to discover the causal relations for our Causal-TGAN. Specifically, we use the package pcalg\footnote{Codes are available at \url{https://cran.r-project.org/web/packages/pcalg/index.html }} to implement causal discovery. In pcalg, it needs a correlation matrix as input for testing conditional independence among variables. However, for tabular data, it is impossible to obtain correlation matrix between continuous and categorical columns. We hence employ the kernel-based approach introduced by \cite{hanEtal19} to get a surrogate correlation matrix of which the correlation between two columns is represented by the distance of the two columns in feature space. It is worthy noting that, we do not claim any novelty in causal discovery. We encourage readers to try different causal discovery methods since accurate causal graphs can reduce the discrepancy between the learned and the target distribution (\cite{gouEtal18}). 

\vspace{-0.2cm}
\subsection{Modeling SCMs with Causal-TGAN}
\vspace{-0.2cm}
\label{sec:causal_TGAN}
In this section, we introduce the construction of Causal-TGAN as an SCM by linking the components of Causal-TGAN with the concepts of SCM introduced in Section \ref{sec:preliminary}. For the triplet representation of an SCM, $ \mathcal M = \langle X, \mathcal F, \mathcal U \rangle$, Causal-TGAN employs the Multiple Layer Perceptron (MLP) for all causal mechanisms in $\mathcal F$. For the dataset with the feature set $\mathcal A$, each endogenous variable $X_i$ represents a feature in $\mathcal A$. The exogenous variables $U$ for every node $X$ are identically and independently sampled from the distribution $\mathcal U$, which is the Gaussian Distribution in our experiment. Given a randomly sampled exogenous variable set $U$ and the causal graph, Causal-TGAN can hence generate a sample $S$ as:
\begin{equation}
    S = \mathcal F(U) = [F_1(PA_1, U_1), F_2(PA_2, U_2),...,F_d(PA_d, U_d)]
\end{equation} 
where d is the number of features and $PA_i \in X \setminus X_i$ denotes the parents nodes of the $i^{th}$ node. Causal-TGAN generates each endogenous variable in the topological order of the causal graph (i.e., from the top nodes to the bottom nodes). Note that, since our causal graph is acyclic, $PA_i$ must be a null set if the $i^{th}$ node is the top node. Consequently, like other GANs, our Causal-TGAN can generate a sample with only noises that sampled from the Gaussian Distribution.

The Causal-TGAN is trained adversarially to approximate the target distribution. We follow the GAN training procedure proposed by \cite{gulEtal17}. We use $\mathcal F^{\theta}$ to denote the data generator, where $\theta$ is the set of parameters for each causal mechanism $F_i^{\theta_{i}}$: $\theta = \{\theta_{1}, \theta_{2},..., \theta_{d}\}$. Let $D^\phi$ denotes the discriminator which is parameterized by $\phi$. We train Causal-TGAN by solving the following optimization problem:
\begin{equation}
    \min_{\theta}\max_{\phi}V(\theta, \phi) = \mathbb{E}_{\boldsymbol{x} \sim p_{data}(\boldsymbol{x})}[\log D^\phi(x)] +  \mathbb{E}_{U \sim \mathcal U}[\log(1-D^\phi(\mathcal F^{\theta}(U)))]
\end{equation}

\section{Experiment}


\subsection{Dataset \& Metrics}
\vspace{-0.2cm}
We use both simulated (causal graph is known) and real datasets (causal graph is unknown) in our experiment. For the simulated dataset, we use four popular Bayesian Networks, which are asia, child, alarm, and insurance\footnote{The structures of Bayesian Networks are available at \url{http://www.bnlearn.com/bnrepository/.}}, to randomly sample 10,000 data points for each network. For the real dataset, we use adult, census, and news datasets from the UCI machine learning repository (\cite{Dua:2019}). In the following section, we use target dataset to denote the dataset used for training. 

Generally, we evaluate the quality of the synthetic data by measuring the \textbf{reality} and the \textbf{machine learning efficacy} for the synthetic data. \textbf{Reality} measures how real the synthetic data compared with the target data. We use different methods to measure the reality for the simulated dataset and real dataset. We regard the Bayesian Networks used for generating the simulated dataset as \textit{oracle model}. For the simulated dataset, we take advantage of the oracle model to calculate the likelihood of the synthetic data on oracle model. As for the real dataset, we use Kolmogorov-Smirnov (KS) Test (for continuous columns) and Chi-Squared (CS) test (for discrete columns) to compare the distribution between the real and synthetic columns. The scores returned by the CS and KS tests indicate the distribution difference. We use $ 1 $ to subtract the returned score so that a value closer to 1 represents a higher degree of similarity. To measure the \textbf{Machine learning efficacy}, we train machine learning models on the synthetic data and then evaluate the trained models on the target dataset. We use classification accuracy for classification tasks and $R^2$ for the regression task to denote the machine learning efficacy score. The used machine learning models for classification are the MLP classifier, AdaBoost, and Decision Tree.
For regression, we use Linear regressor and MLP regressor. 
We use implementations provided by SDV package\footnote{Codes with examples are available at \url{https://sdv.dev/SDV/user_guides/evaluation/single_table_metrics.html}} for all the metrics above.

\subsection{Results}
\vspace{-0.2cm}
\begin{table}[t!]
\centering
\footnotesize
\scalebox{1}{
\begin{tabular}{cccccc}
\toprule 
Method & asia & child & alarm & insurance & Average \\
\toprule
Identity & -2.24 & -12.15 & -10.42 & -13.04 & -9.46\\ 
\midrule
PrivBN & \underline{-2.29} & \textbf{-12.39} & \underline{-11.90} & \underline{-15.03} & \underline{-10.40}\\ 
MedGAN &-2.81 & -14.23 & -19.31 & -17.16 & -13.37\\ 
TableGAN & -3.70 & -15.47 & -25.36 & -16.42 & -15.23\\ 
CTGAN &-2.56 & -14.40 &-19.79 & -17.00& -13.43 \\ 
Causal-TGAN (Ours) & \textbf{-2.26} & \underline{-12.42} & \textbf{-11.28} & \textbf{-13.53} & \textbf{-9.87}\\
\bottomrule
\end{tabular}}
\caption{Log-Likelihood of Synthetic Simulated Data on the Oracle Models. Higher value indicates higher likelihood the synthetic data is generated from the same origin as the target data.}
\label{tab:simulated}
\end{table}

We compare our Causal-TGAN with PrivBN (\cite{zhaEtal17}), MedGAN (\cite{choEtal17}), TableGAN (\cite{Park18}) and CTGAN (\cite{Xu19}). We train each model for 300 epochs with batch size of 500. We follow the method proposed in \cite{Xu19} to encode the variables. Specifically, we use Gaussian Mixture Model to encode continuous columns and one-hot to encode discrete columns. For the below tables, we consistently highlight the best performance with bold and underline the second best performance. Moreover, we denote Identity as the method that generate target dataset.

\begin{table}[h!]
\centering
\footnotesize
\scalebox{1}{
\begin{tabular}{cccccccc}
\toprule 
\multirow{2}{*}{Method}& \multicolumn{2}{c}{adult} & \multicolumn{2}{c}{census} & \multicolumn{2}{c}{news} & \multirow{2}{*}{Average} \\ \cmidrule{2-7}
&CS Test & KS Test & CS Test & KS Test &CS Test & KS Test &\\
\toprule
Identity & 1.00 & 1.00 & 1.00 & 1.00 & 1.00 & 1.00 & 1.00\\ 
\midrule
PrivBN & \textbf{0.91} & 0.37 & \textbf{0.99} & \textbf{0.94} & - & 0.59 & 0.76\\ 
MedGAN & 0.68 & 0.04 & 0.79 & 0.12 & - & \textbf{1.00} & 0.53\\ 
TableGAN & \underline{0.88} & 0.41 & 0.79 & 0.64 & - & 0.39 & 0.62\\ 
CTGAN & 0.87 & \underline{0.79} & \underline{0.95} & 0.79& - & \underline{0.86} & \textbf{0.85} \\ 
Causal-TGAN (Ours) & \underline{0.88} & \textbf{0.83} &  0.93 & \underline{0.83} & - &  0.79 & \textbf{0.85}\\
\bottomrule
\end{tabular}}
\caption{Distribution similarity between columns from synthetic dataset and target dataset. CS Test for discrete columns and KS Test for continuous columns. }
\label{tab:real}
\end{table}

\begin{wraptable}{r}{5.5cm}
\centering
\footnotesize
\scalebox{0.7}{
\begin{tabular}{cccc}
\toprule 
\multirow{2}{*}{Method} & \multicolumn{2}{c}{Classification} & Regression \\ \cmidrule{2-4} 
& adult & census & news  \\
\toprule
Identity & 0.885 & 0.938 & 0.014 \\ 
\midrule
PrivBN & 0.837 & \underline{0.937} & -4.49\\ 
MedGAN & 0.840 & \textbf{0.968} & -8.80\\ 
TableGAN & 0.730 & 0.708 & -3.09\\ 
CTGAN & \textbf{0.872} & 0.928 & \textbf{-0.43}  \\ 
Causal-TGAN (Ours) & \underline{0.847} & 0.903 & \underline{-0.53}\\
\bottomrule
\end{tabular}}
\caption{\textbf{Machine Learning Efficacy Score} The reportd score are average accuracy for classification tasks and average $R^2$ score for regression task.}
\label{tab:ml}
\end{wraptable}

For the simulated data, we report the log-likelihood of synthetic data on the oracle Bayesian Network in Table \ref{tab:simulated}. Causal-TGAN achieves the best overall performance as well as the best for three datasets except for the child dataset. Moreover, our Causal-TGAN outperforms other GAN models by an average of $29\%$. In this experiment, we use the structure of the Bayesian Network as the causal graph for Causal-TGAN. This result demonstrates an ideal case: when the true underlying causal relations are presented, causality-based GAN models can outperform those that capture correlation.

For the real dataset, we first measure the distribution similarity between the columns of the real and synthetic dataset, see Table \ref{tab:real}. The scores vary for different models across the different datasets. We hence compare by the average scores over all the dataset. It shows that both CTGAN and Causal-TGAN achieve the best average column similarity. All columns of the news dataset (binary and continuous) are considered continuous columns in our experiment. As a result, there is no CS Test score reported for the news dataset. As for the machine learning efficacy, Table \ref{tab:ml} reports the accuracy and $R^2$ scores for classification and regression tasks, respectively. For adult and news data, CTGAN keeps achieving the best performance for all datasets, and Causal-TGAN consistently achieves the second-best performance. For census, MedGAN and PrivBN achieve the first two best scores.

\vspace{-0.2cm}
\section{Conclusion}
\vspace{-0.3cm}
We propose a novel generative model named Causal-TGAN to generate tabular data by taking advantage of inter-variants causal relations. We evaluate Causal-TGAN with metrics of distribution similarity and machine learning efficacy. The experimental results show that our Causal-TGAN with the presence of true causal relations can outperform all other generative models in generating realistic synthetic data. When the true causal relations are absent, Causal-TGAN can still achieve comparative results by using the estimated causal relations. In future work, we will discover more advantages of using causal relations for generating tasks.

\bibliography{iclr2021_conference}

\begin{thebibliography}{15}
\providecommand{\natexlab}[1]{#1}
\providecommand{\url}[1]{\texttt{#1}}
\expandafter\ifx\csname urlstyle\endcsname\relax
  \providecommand{\doi}[1]{doi: #1}\else
  \providecommand{\doi}{doi: \begingroup \urlstyle{rm}\Url}\fi

\bibitem[Camino et~al.(2018)Camino, Hammerschmidt, and State]{Camino18}
Ramiro~D. Camino, Christian~A. Hammerschmidt, and Radu State.
\newblock Generating multi-categorical samples with generative adversarial
  networks.
\newblock In \emph{ICML workshop on Theoretical Foundations and Applications of
  Deep Generative Models}, 2018.

\bibitem[Choi et~al.(2017)Choi, Biswal, Malin, Duke, Stewart, and
  Sun]{choEtal17}
Edward Choi, Siddharth Biswal, Bradley Malin, Jon Duke, Walter~F Stewart, and
  Jimeng Sun.
\newblock Generating multi-label discrete patient records using generative
  adversarial networks.
\newblock In \emph{Machine learning for healthcare conference}, pp.\  286--305.
  PMLR, 2017.

\bibitem[Dua \& Graff(2017)Dua and Graff]{Dua:2019}
Dheeru Dua and Casey Graff.
\newblock {UCI} machine learning repository, 2017.
\newblock URL \url{http://archive.ics.uci.edu/ml}.

\bibitem[Goodfellow et~al.(2014)Goodfellow, Pouget-Abadie, Mirza, Xu,
  Warde-Farley, Ozair, Courville, and Bengio]{Goodfellow14}
Ian Goodfellow, Jean Pouget-Abadie, Mehdi Mirza, Bing Xu, David Warde-Farley,
  Sherjil Ozair, Aaron Courville, and Yoshua Bengio.
\newblock Generative adversarial networks.
\newblock In \emph{Generative Adversarial Nets}, pp.\  2672--2680, 2014.

\bibitem[Goudet et~al.(2018)Goudet, Kalainathan, Caillou, Guyon, Lopez-Paz, and
  Sebag]{gouEtal18}
Olivier Goudet, Diviyan Kalainathan, Philippe Caillou, Isabelle Guyon, David
  Lopez-Paz, and Michele Sebag.
\newblock Learning functional causal models with generative neural networks.
\newblock In \emph{Explainable and interpretable models in computer vision and
  machine learning}, pp.\  39--80. Springer, 2018.

\bibitem[Gulrajani et~al.(2017)Gulrajani, Ahmed, Arjovsky, Dumoulin, and
  Courville]{gulEtal17}
Ishaan Gulrajani, Faruk Ahmed, Martin Arjovsky, Vincent Dumoulin, and Aaron
  Courville.
\newblock Improved training of wasserstein gans.
\newblock \emph{arXiv preprint arXiv:1704.00028}, 2017.

\bibitem[Handhayani \& Cussens(2019)Handhayani and Cussens]{hanEtal19}
Teny Handhayani and James Cussens.
\newblock Kernel-based approach to handle mixed data for inferring causal
  graphs.
\newblock \emph{arXiv preprint arXiv:1910.03055}, 2019.

\bibitem[Jordon et~al.(2019)Jordon, Yoon, and Schaar]{Jordon19}
James Jordon, Jinsung Yoon, and Mihaela van~der Schaar.
\newblock Pate-gan: Generating synthetic data with differential privacy
  guarantees.
\newblock In \emph{International Conference on Learning Representations}, 2019.

\bibitem[Kocaoglu et~al.(2017)Kocaoglu, Snyder, Dimakis, and
  Vishwanath]{kocEtal17}
Murat Kocaoglu, Christopher Snyder, Alexandros~G Dimakis, and Sriram
  Vishwanath.
\newblock Causalgan: Learning causal implicit generative models with
  adversarial training.
\newblock \emph{arXiv preprint arXiv:1709.02023}, 2017.

\bibitem[Park et~al.(2018)Park, Mohammadi, Gorde, Jajodia, Park, and
  Kim]{Park18}
Noseong Park, Mahmoud Mohammadi, Kshitij Gorde, Sushil Jajodia, Hongkyu Park,
  and Youngmin Kim.
\newblock Data synthesis based on generative adversarial networks.
\newblock In \emph{The 44th International Conference on Very Large Data Bases},
  volume~11, 2018.

\bibitem[Pearl(2009)]{pearl09}
Judea Pearl.
\newblock \emph{Causality}.
\newblock Cambridge university press, 2009.

\bibitem[Spirtes et~al.(2000)Spirtes, Glymour, Scheines, and Heckerman]{pc00}
Peter Spirtes, Clark~N Glymour, Richard Scheines, and David Heckerman.
\newblock \emph{Causation, prediction, and search}.
\newblock MIT press, 2000.

\bibitem[Torfi et~al.(2020)Torfi, , and Fox]{Torfi20}
Amirsina Torfi, , and Edward~A. Fox.
\newblock Corgan: Correlation-capturing convolutional generative adversarial
  networks for generating synthetic healthcare records.
\newblock In \emph{The 33rd International FLAIRS Conference, AI in Healthcare
  Informatics}, pp.\  335--340, 2020.

\bibitem[Xu et~al.(2019)Xu, Skoularidou, Cuesta-Infante, and
  Veeramachaneni]{Xu19}
Lei Xu, Maria Skoularidou, Alfredo Cuesta-Infante, and Kalyan Veeramachaneni.
\newblock Modeling tabular data using conditional gan.
\newblock In \emph{33rd Conference on Neural Information Processing Systems},
  2019.

\bibitem[Zhang et~al.(2017)Zhang, Cormode, Procopiuc, Srivastava, and
  Xiao]{zhaEtal17}
Jun Zhang, Graham Cormode, Cecilia~M Procopiuc, Divesh Srivastava, and Xiaokui
  Xiao.
\newblock Privbayes: Private data release via bayesian networks.
\newblock \emph{ACM Transactions on Database Systems (TODS)}, 42\penalty0
  (4):\penalty0 1--41, 2017.

\end{thebibliography}
\bibliographystyle{iclr2021_conference}


\end{document}